\documentclass{article}

\usepackage[preprint]{neurips_2026}

\usepackage[utf8]{inputenc} 
\usepackage[T1]{fontenc}    
\usepackage{hyperref}       
\usepackage{url}            
\usepackage{booktabs}       
\usepackage{amsfonts}       
\usepackage{nicefrac}       
\usepackage{microtype}      
\usepackage{xcolor}         

\usepackage{amssymb}
\usepackage{pifont}
\usepackage{natbib}

\usepackage{amsmath}
\usepackage{graphicx}

\usepackage{booktabs}
\usepackage{multirow}
\usepackage{enumitem}
\usepackage{subcaption}

\usepackage{makecell}

\newcommand{\xmark}{\ding{55}}
\title{Real vs. Semi-Simulated: Rethinking Evaluation for Treatment Effect Estimation}

\author{%
  George Panagopoulos\\
  Department of Computer Science\\
  University of Luxembourg\\
  2 Av. de l'Universite, 4365 Belval Esch-sur-Alzette\\
  \texttt{georgios.panagopoulos@uni.lu} \\
}

\begin{document}

\maketitle

\begin{abstract}
Estimating heterogeneous treatment effects with machine learning has attracted substantial attention in both academic research and industrial practice. However, the two communities often evaluate models under markedly different conditions. Methodological work typically relies on semi-simulated benchmarks and metrics that require counterfactual outcomes, whereas real-world applications rely on observable metrics based on ranking or test outcomes. Despite the well-known gap between methodological progress and practical deployment, the relationship between these evaluation regimes has not been examined systematically.
We conduct a large-scale empirical study of treatment effect evaluation across standard semi-simulated benchmark families and real-world datasets. Our benchmark covers meta-learners paired with multiple base learners, as well as specialized causal machine learning models. We evaluate these methods using observable metrics common in application-oriented literature, alongside counterfactual metrics commonly used in methods papers.
Our results reveal two complementary gaps. First, counterfactual metrics do not reliably recover the estimators preferred by observable metrics, even on the same semi-simulated benchmarks. Second, rankings obtained on semi-simulated benchmarks do not transfer to real datasets. We further find that simple meta-learners with strong base models are consistently competitive, in contrast to specialized causal models.
Overall, our findings suggest that progress in treatment effect estimation research should not be assessed solely through counterfactual metrics and semi-simulated benchmarks, but it would benefit from incorporating observable metrics and real-data validation.
\end{abstract}


\section{Introduction}
Machine learning for causal inference, often referred to as causal machine learning (\textbf{Causal ML}), has made significant strides in recent years, proving vital for fields such as economics, marketing, and health \citep{bica2021real,betlei2021uplift,jaskowski2012uplift}. While a plethora of methods have been developed \citep{shalit2017estimating,kunzel2019metalearners,shi2019adapting,melnychuk2022causal,javaloy2023causal}, their adoption in industrial use cases has remained limited \citep{feuerriegel2024causal,curth2021really}. One of the core reasons for this hesitation is the lack of solid evaluation frameworks that abide by real-world practices. 
Specifically, evaluation data and metrics designed for estimating Conditional Average Treatment Effect (\textbf{CATE}) or Individual Treatment Effect (\textbf{ITE)} , suffer an inherent disadvantage: they require predicting the outcome of a sample under a counterfactual setting, i.e., what happens when they are treated versus when they are not. Unlike standard machine learning, the counterfactual outcome is unobserved by definition, and hence when a model predicts an effect $Y_i(1)-Y_i(0)$, only one of those potential outcomes is ever available for validation.

This has led to two distinct evaluation paths. The first practice, dominant in causal ML papers, evaluates models on synthetic or semi-simulated benchmarks where both potential outcomes are generated by construction. 
However, semi-simulated benchmarks can encode data-generating assumptions \citep{curth2021really,poinsotposition}, while surrogate evaluation criteria may systematically favor some estimators \citep{curth2023search,mahajan2022empirical}.
The second practice, common in industrial applications, evaluates models using observable quantities, such as ranking metrics or factual test outcomes \citep{moraes2023uplift}. These metrics do not measure individual counterfactual error directly, but they are closer to real deployment goals.
Thus, a gap has emerged between research and real-world evaluation. 
If the two are misaligned, research progress may be distorted by rewarding methods that perform well under artificial evaluation regimes but do not transfer to deployment-oriented settings.
We thus aim to study the difference between research practice and real-world evaluation by answering this question: if a practitioner selects a causal ML model based on counterfactual metrics and semi-simulated benchmarks, how will it perform on observable metrics and real data?

We conduct, to the best of our knowledge, the largest empirical comparison of treatment effect evaluation regimes to date, covering 117 benchmark instances (112 semi-simulated and 5 real-world datasets), with sample sizes ranging from hundreds to nearly 14 million. Our estimator pool spans 5 standard meta-learners \citep{kunzel2019metalearners} with 3 variations of base-learners as well as neural causal architectures \citep{shalit2017estimating,shi2019adapting} and Causal Forests \citep{wager2018estimation}. 
Across these methods, we compare three families of metrics: counterfactual-based metrics, observable outcome metrics, and observable ranking metrics, as analyzed in \ref{sec:metrics}. The overall conclusions can be summarized as follows:

\begin{itemize}
    \item \textbf{Metric Disconnect:} We find that counterfactual metrics on semi-simulated data do not reliably recover the estimators preferred by observable metrics. In fact, models selected via the academic research regime often incur substantial regret when evaluated on observable metrics, even on the same semi-simulated benchmark.
    \item \textbf{Dataset Non-Transferability:} Rankings obtained on semi-simulated benchmarks do not transfer to real datasets, even when using identical observable metrics. Moreover, in a matched case study between an industrial-scale dataset and its semi-simulated variant, the overall ordering and top-ranked estimators are not fully preserved, despite identical covariates and treatment assignments.
    \item \textbf{Practical Conclusions:} We demonstrate that progress in the field should not be assessed through oracle metrics alone, but rather incorporating observable metrics and real-data validation is essential to close the gap between research and deployment. Secondly, we observe that simple meta-learners with strong predictive backbones are consistently competitive across most regimes, supporting their inclusion as benchmarks in methodological papers. 
\end{itemize}

\section{Related Work}
Several recent works have examined evaluation and model selection for treatment effect estimation. 
\cite{gutierrez2017causal} connect uplift modeling with the potential-outcomes framework and reviews ranking-oriented evaluation tools such as uplift and Qini curves. 
\cite{schuler2018comparison} compare validation criteria for ITE model selection and find that $\tau$-risk-based objectives perform well, although their study is restricted to simulated data-generating processes. 
\cite{olaya2020survey} benchmark multitreatment uplift methods on real datasets using Qini and expected-response metrics. 
\cite{saito2020counterfactual} similarly focus on stable CATE model selection, proposing counterfactual cross-validation to better preserve the rank ordering of candidate estimators.
\cite{curth2021really} show that common semi-synthetic CATE benchmarks can favor particular combinations of data-generating mechanisms and estimators. 
\cite{curth2023search} analyze surrogate model-selection criteria in controlled simulations, showing that factual prediction criteria can be poorly targeted for CATE selection and that plug-in criteria can suffer from congeniality bias. 
\cite{rafla2022evaluation} study how non-random assignment bias affects uplift-model evaluation, benchmarking several uplift methods on real and synthetic datasets using Qini. 
More recently, \cite{poinsotposition} argued that synthetic experiments remain necessary for causal machine learning because the causal ground truth is observable and assumptions can be varied in isolation. However, they emphasize that their design choices must be made explicit and stress-tested.  
Closest to our work in scale, \cite{mahajan2022empirical} benchmark surrogate model-selection criteria across many estimators and datasets, but evaluate success primarily through Precision in Estimation of Heterogeneous Effect (\textbf{PEHE}) on semi-simulated or synthetic benchmarks.

Table~\ref{tab:metric_comparison} summarizes the scope of these studies. 
Our work differs primarily in the question it asks: rather than proposing or selecting a surrogate criterion for PEHE, we ask whether the evaluation regimes commonly used in methods papers align with the objectives used in real-world applications. 
This requires jointly comparing counterfactual and observable metrics, which include outcome and ranking-based metrics, across both semi-simulated and real datasets. 
In this sense, our study is complementary to prior work, which primarily evaluates how well candidate criteria select estimators under oracle targets, PEHE-based optimal metrics, or controlled experimental knobs, whereas we test whether the majority of these metrics are a reliable proxy for real-data evaluation.


\begin{table}[t]
\centering
\scriptsize
\setlength{\tabcolsep}{3pt}
\begin{tabular}{lccccccc}
\toprule
\textbf{Work} 
& \textbf{Sim} 
& \textbf{Real} 
& \textbf{CF}  
& \textbf{Outcome} 
& \textbf{Ranking} 
& \textbf{Data No.} 
& \textbf{Max $n$}  \\
\midrule

\cite{gutierrez2017causal} 
& \checkmark & \xmark & \xmark & \checkmark & \checkmark 
& 1 & 10{,}000  \\ 

\cite{schuler2018comparison} 
& \checkmark & \xmark & \checkmark & \checkmark & \xmark 
& 16 & 3{,}000 \\

\cite{olaya2020survey} & \xmark & \checkmark & \xmark & \checkmark & \checkmark & 8 & 180{,}002 \\

\cite{saito2020counterfactual} 
& \checkmark & \xmark & \checkmark & \checkmark & \xmark 
& 1 & 747  \\

\cite{curth2021really} 
& \checkmark & \xmark & \checkmark & \xmark & \xmark 
& 2 & 4{,}802  \\

\cite{curth2023search} 
& \checkmark & \xmark & \checkmark & \checkmark & \xmark 
& 3 & 4{,}802 \\

\cite{rafla2022evaluation} & \checkmark & \checkmark & \xmark & \xmark & \checkmark & 8 & 600{,}000 \\
\cite{poinsotposition} 
& \checkmark & \xmark & \checkmark & \xmark & \xmark 
& 2 & --\\

\cite{mahajan2022empirical} 
& \checkmark & \xmark & \checkmark & \checkmark & \checkmark 
& 78 & 14{,}559\\
\midrule
\textbf{This work} 
& \checkmark & \checkmark & \checkmark & \checkmark & \checkmark 
& \textbf{117} & \textbf{13{,}979{,}592} \\
\bottomrule
\end{tabular}
\caption{Comparison with related evaluation studies. 
``Sim'' denotes simulated or semi-simulated data; ``Real'' denotes real datasets; ``CF/Ranking/Outcome'' denotes counterfactual/ranking/outcome-based metrics respectively as defined in sec \ref{sec:metrics}; ``Data No.'' denotes the number of datasets or benchmark instances; and ``Max $n$'' denotes the largest dataset size.} 
\label{tab:metric_comparison}
\end{table}


\section{Experimental Design}

\subsection{Problem Setting}
\label{sec:problem_setting}

We consider binary treatment-effect estimation in the potential-outcomes
framework. For each unit, we observe covariates $X \in \mathcal{X}$, a treatment
indicator $T \in \{0,1\}$, and an outcome $Y$. The two potential outcomes
$Y(0)$ and $Y(1)$ denote the outcomes that would be realized under control and
treatment, respectively. Since each unit receives only one treatment, the
observed outcome is $Y_i = Y_i(T_i)$.
The target estimand is the conditional average treatment effect (CATE),
\[
\tau(x)
=
\mathbb{E}\!\left[Y(1)-Y(0)\mid X=x\right],
\]
which can also be written as the difference between the two potential-outcome
regression functions, with
$\mu_t(x)=\mathbb{E}[Y(t)\mid X=x]$ for $t\in\{0,1\}$.
For the causal interpretation of this estimand, we rely on the standard
identification conditions used in treatment-effect estimation: consistency,
conditional exchangeability, and overlap \citep{rubin2005causal}. Consistency links the observed outcome
to the potential outcome under the treatment actually received. Conditional
exchangeability requires that, after conditioning on $X$, treatment assignment
does not depend on the potential outcomes. Overlap requires that both treatment
arms have a positive probability for the covariate regions under study. Under
these conditions, the potential outcome regression is identified from observed
data 
\[
\mu_t(x)
=
\mathbb{E}[Y \mid T=t, X=x],
\qquad t\in\{0,1\},
\]
and therefore the CATE can be estimated from the contrast between the two
observed conditional outcome regressions $\mu_1(x)-\mu_0(x)$.

Although CATE can be identifiable at the population level, the
individual treatment effect $Y_i(1)-Y_i(0)$ is not observed for real data,
because one of the two potential outcomes is missing for every unit. As a
result, real datasets do not provide a supervised label against which a CATE
prediction can be directly validated. This creates a model-selection problem:
Given a set of candidate estimators
$\mathcal{A}=\{a_1,\ldots,a_K\}$, where each estimator outputs
$\hat{\tau}_a(x)$, we want to determine which estimator should be preferred for
downstream use.
Semi-simulated benchmarks and real datasets differ precisely in what they allow
us to evaluate. In semi-simulated benchmarks, both potential outcomes are
generated by construction, so counterfactual metrics can be computed. In real
datasets, evaluation must instead rely on quantities that are observable from
the test data. We therefore compare two metric families: observable vs counterfactual-based. Observables include ranking metrics,
which assess treatment prioritization, and outcome-based metrics, which use observed
test outcomes. Counterfactual metrics require simulated potential
outcomes. Formally, let $M$ denote an evaluation metric and $D$ a dataset. Each metric
assigns a score $m_M(a;D)$ to an estimator $a$. After orienting all metrics so
that lower scores indicate better performance, we convert scores into
metric-induced ranks,
\[
R_M(a;D)
=
\operatorname{rank}_{a'\in\mathcal{A}}
\bigl(m_M(a';D)\bigr),
\]
where lower rank is better. Our goal is to test whether different evaluation
regimes lead to consistent model rankings. This allows us to quantify the gap
between research-oriented benchmarking practices and deployment-oriented
evaluation criteria. Specifically, we address the following questions:
\begin{enumerate}[label=\textbf{Q\arabic*.}]
    \item \textbf{Metric transfer.}
    \textit{Do counterfactual metrics computed on semi-simulated benchmarks
    induce the same model rankings as observable metrics?}

    \item \textbf{Dataset transfer.}
    \textit{Do model rankings obtained with observable metrics on
    semi-simulated benchmarks transfer to observable-metric rankings on real
    datasets?}
\end{enumerate}
We also report several secondary findings and practical takeaways from the
benchmarking study.

\subsection{Metrics}
\label{sec:metrics}

We evaluate each estimator using observable and counterfactual metrics. Full definitions are given in Appendix~\ref{app:metric_definitions}.

\paragraph{Observable outcome-based metrics.}
Outcome-based metrics use only observed test outcomes. We use factual loss \citep{curth2023search}, which
measures prediction error under the observed treatment assignment, and
$\Delta\mathrm{ATE}$ \citep{curth2021really}, which measures the discrepancy
between the predicted treatment effect and an empirical average treatment effect
estimate from the test data. We use $\Delta\mathrm{ATE}$ as an observable aggregate diagnostic rather than
as an oracle measure of treatment-effect accuracy. 

\paragraph{Observable ranking-based metrics.}
Ranking metrics evaluate whether an estimator correctly prioritizes units for
treatment. We consider three standard ranking-oriented criteria: Up@20
\citep{rafla2022evaluation}, which measures the treatment effect among the top
20\% of units ranked by the model; the Qini score \citep{diemert2018large},
which measures the gain from model-based targeting relative to random targeting;
and RATE AUTOC \citep{yadlowsky2025evaluating}, which summarizes the targeting
operator characteristic curve with higher weight on top-ranked units. It should be noted that ranking metrics are not treated as an optimal evaluation strategy, as their limitations are well known \citep{zhu2025rethinking}. Rather, we use them as observable indicators of deployment-oriented performance.

\paragraph{Counterfactual-based metrics.}
Counterfactual metrics require access to both potential outcomes and are
therefore restricted to simulated or semi-simulated settings. 
We use
$\sqrt{\mathrm{PEHE}}$ \citep{alaa2017bayesian}, which measures individual-level
treatment-effect error; policy value \citep{curth2021really}, which evaluates
the expected outcome under the policy induced by the estimated effects; and
$\Delta\mathrm{ATE}(\mathrm{Sim})$ \citep{johansson2016learning}, which compares the estimated ATE with the true simulated ATE. We do not examine plug-in surrogate metrics, as they have been extensively
studied in prior work \citep{mahajan2022empirical,curth2023search} and are known to suffer from issues such as congeniality bias.

\subsection{Data}
Table (\ref{tab:dataset_summary}) summarizes the datasets. The semi-simulated data splits are predefined by the \textit{CATENets} package \citep{curth2021nonparametric} for \textbf{IHDP} \citep{hill2011bayesian} and \textbf{Twins} \citep{louizos2017causal}, and \textit{causallib} \citep{shimoni2019evaluation} for \textbf{ACIC2016} \citep{dorie2019automated}. 
In contrast, \textbf{Hillstrom Men/Women} \citep{hillstrom2008minethatdata}, \textbf{NHEFS} \citep{hernan2010causal}, \textbf{Retail} \citep{rafla2022evaluation}, \textbf{Criteo} \citep{diemert2018large} and \textbf{Criteo-ITE} \citep{diemert2021large} follow a 5-fold cross-validation. Retail undergoes the same preprocessing as in \cite{panagopoulos2024uplift}. 
The metrics are averaged over all simulations / folds and over 5 experiments with different random seeds.

\begin{table}[t]
\centering
\small
\begin{tabular}{lccc lcc}
\toprule
\multicolumn{4}{c}{\textbf{Semi-simulated}} 
& \multicolumn{3}{c}{\textbf{Real}} \\
\cmidrule(lr){1-4} \cmidrule(lr){5-7}
\textbf{Dataset} & \textbf{\# Samples} & \textbf{\# Features} & \textbf{\# Versions}
& \textbf{Dataset} & \textbf{\# Samples} & \textbf{\# Features} \\
\midrule
\textbf{IHDP}       & 747        & 25 & 100 
& \textbf{NHEFS}         & 1,566      & 18 \\
\textbf{ACIC2016}   & 4,802      & 79 & 10  
& \textbf{Hillstrom M/W} & 42,613     & 12  \\
\textbf{Twins}      & 11,400     & 39 & 1   
& \textbf{Retail}        & 180,653    & 7   \\
\textbf{Criteo-ITE}& 13,979,592 & 12 & 1   
& \textbf{Criteo}        & 13,979,592 & 12 \\
\bottomrule
\end{tabular}
\caption{Summary of datasets used in the benchmark.}
\label{tab:dataset_summary}
\end{table}

\subsection{Methods}

We evaluate S-, T-, X- \citep{kunzel2019metalearners}, R- \citep{nie2021quasi} learners and the Doubly Robust (DR) model \citep{chernozhukov2018double} with three base learners: linear/logistic regression (LR), XGBoost (XGB), and two-layer MLPs (NN). When a propensity model is required, it is fit with an ElasticNet regression. We additionally include TARNet, CFR \citep{shalit2017estimating}, DragonNet \citep{shi2019adapting}, and Causal Forest \citep{wager2018estimation} as commonly used specialized baselines. 
The meta-learners undergo hyperparameter optimization, and their ranges are described in Appendix \ref{app:hyperparameters}, along with more concrete definitions of the methods. 
We use the standard \textit{EconML} \citep{econml} implementations of the meta-learners. For meta-learners such as the X-learner, R-learner, and DR, nuisance models are fit within the training fold. Since certain metrics such as factual loss cannot be computed in models that do not produce individual outcome predictions, i.e., direct learners like  X-, R-, DR-learner, we use the outcome models that are created as part of their computation as an approximation. If this is not present, e.g., for Causal Forest, the result is left blank.  

%

\section{Results}
We report compact summaries of the ranking patterns. Full average-rank heatmaps  are provided in Appendix~\ref{app:rank_heatmaps}, and the detailed
metric values are reported in Appendix~\ref{app:detailed_results}. Unless
otherwise stated, ranks are computed after averaging metric values over folds,
simulations, and random seeds. Any exact ties are explicitly mentioned in the tables.

\subsection{Q1: Different metric families induce different model rankings}
\label{sec:results_metric_dataset_disagreement}

Table~\ref{tab:simulated_top_methods_all} reports the top-ranked estimator under each metric for the semi-simulated datasets. The main pattern is immediate: the ``best'' method is highly metric-dependent. On \textbf{ACIC2016}, for example, the ranking metrics select three different XGBoost-based meta-learners.
In contrast, all three counterfactual-based metrics select X(XGB) and outcome metrics indicate  Causal Forest and T(XGB). Thus, even within the same semi-simulated benchmark, the estimator preferred by counterfactual metrics is not necessarily the estimator preferred by observable metrics. This disagreement is not specific to \textbf{ACIC2016}. On \textbf{IHDP}, ranking metrics favor DR(LR) or S(XGB), whereas counterfactual and outcome metrics favor DragonNet or T(XGB), while the disagreement persists on \textbf{Twins} where it is almost ubiquitous.

Table~\ref{tab:selection_regret} quantifies the practical cost of selecting models using counterfactual metrics. We report the average rank, under each observable metric, of the estimator selected by the counterfactual metrics. The resulting regret is substantial. In particular, the estimator selected by $\sqrt{\mathrm{PEHE}}$ ranks poorly under \textsc{Uplift@20}, Qini, and RATE, with average ranks of $13$, $12.3$, and $10.3$, respectively. Policy value and $\Delta\mathrm{ATE}(\mathrm{Sim})$ are closer to the observable rankings but still incur a regret of at least  $6.7$ ranks. The only exception is factual loss, which appears more aligned with the counterfactual criteria. However, factual loss itself has several disadvantages on treatment-effect estimation evaluation. Specifically, it is known to be affected by the covariate shift between treatment groups and can favor nuisance prediction rather than accurate CATE estimation \citep{curth2023search}.  

\begin{table}[t]
\centering
\small
\begin{tabular}{lrrr}
\toprule
\textbf{Metric} & $\boldsymbol{\sqrt{\mathrm{PEHE}}}$ & \textbf{Policy} & 
$\boldsymbol{\Delta}$\textbf{ATE (Sim)} \\
\midrule
Uplift@20      & 13 & 8.3 & 8.3 \\
Qini           & 12.3 & 6.7 & 6.7 \\
RATE           &  10.3 &     7.5 &                 7.5\\
Factual        &  4.7 & 3.5 & 3.5 \\
$\Delta$ATE    &  7 & 8.3 & 8.3 \\
\bottomrule
\end{tabular}
\caption{Observable rank of counterfactual-selected models. Each entry reports the rank of the model selected as best by a counterfactual metric, under an observable metric. Results are averaged across all semi-simulated datasets. Lower is better. The best method on average for $\boldsymbol{\sqrt{\mathrm{PEHE}}}$ is DragonNet while for \textbf{Policy} and $\textbf{ATE (Sim)}$ it is X (XGN). }
\label{tab:selection_regret}
\end{table}

\begin{table}[h!]
\centering
\small
\begin{tabular}{llccc}
\toprule
\textbf{Type} & \textbf{Metric} & \textbf{ACIC2016} & \textbf{IHDP} & \textbf{Twins} \\
\midrule
\multirow{3}{*}{\textbf{Ranking}} 
& Uplift@20 & T (XGB) & DR (LR) & CFR \\
& Qini      & R (XGB) & DR (LR) & DR (NN) \\
& RATE      & DR (XGB) & S (XGB) & Causal Forest  \\
\midrule
\multirow{2}{*}{\textbf{Outcome}} 
& $\Delta \mathrm{ATE}$ & Causal Forest & DragonNet & R (LR) \\
& Factual               & T (XGB) & DragonNet & S (NN) \\
\midrule
\multirow{3}{*}{\textbf{Counterfactuals}} 
& $\Delta \mathrm{ATE}\,(\mathrm{Sim})$ & X (XGB) & DragonNet & DR (XGB) \\
& Policy                & X (XGB) & T (XGB) & S (LR) \\
& $\sqrt{\mathrm{PEHE}}$ & X (XGB) & DragonNet & R (LR) \\
\bottomrule
\end{tabular}
\caption{Best methods across evaluation metrics for semi-simulated datasets.}
\label{tab:simulated_top_methods_all}
\end{table}

These examples suggest that metric choice can alter the selected estimator, but top-1 comparisons alone do not show whether this disagreement persists across the full ranking of candidate methods. We therefore compare metric-induced rankings across all $111$ instances of the datasets in Table \ref{tab:simulated_top_methods_all} using Kendall's tau and the test described in Appendix~\ref{app:wilcoxon}. The agreement between ranking metrics is significantly higher than their agreement with counterfactual metrics: the mean agreement is $0.355$ among ranking metrics and $0.256$ between ranking and counterfactual metrics, yielding a mean paired difference of $D=0.100$ and a one-sided Wilcoxon signed-rank test of $p=4.5\times 10^{-6}$. This provides statistically significant evidence that deployment-oriented ranking metrics induce rankings that are more mutually aligned with one another than with counterfactual metrics. 

Overall, the results indicate that metric choice can materially affect empirical conclusions. Counterfactual metrics computed on semi-simulated data do not reliably recover the estimators preferred by deployment-oriented ranking metrics, and the same set of estimators can lead to different conclusions depending on the evaluation criterion. Treating $\sqrt{\mathrm{PEHE}}$ or related counterfactual criteria as gold-standard model-selection objectives, as is common in the causal ML literature, can therefore lead to empirical conclusions that are misaligned with deployment-relevant objectives.

\begin{table}[h!]
\centering
\small
\begin{tabular}{llcccc}
\toprule
\textbf{Type} & \textbf{Metric} 
& \textbf{Hillstrom Men} & \textbf{Hillstrom Women} & \textbf{NHEFS} & \textbf{Retail} \\
\midrule
\multirow{3}{*}{\textbf{Ranking}} 
& Uplift@20 & T (LR) & DR (NN) & S (XGB) & Causal Forest \\
& Qini      & S (LR) & X (XGB) & S (XGB) & DR (NN) \\
& RATE      & DR (NN) & DR (NN) & S (XGB) & DR (XGB) \\
\midrule
\multirow{2}{*}{\textbf{Outcome}} 
& $\Delta \mathrm{ATE}$ & T (LR) & T (LR) & CFR & S (XGB) \\
& Factual               & TARNet & CFR & S (LR) & DR (XGB) \\
\bottomrule
\end{tabular}
\caption{Best methods across observable evaluation metrics for the real datasets.}
\label{tab:real_top_methods}
\end{table}

\subsection{Q2: Rankings do not transfer from semi-simulated to real data}
\label{sec:results_data_disagreement}

\begin{figure}[t]
    \centering
    \includegraphics[width=\linewidth]{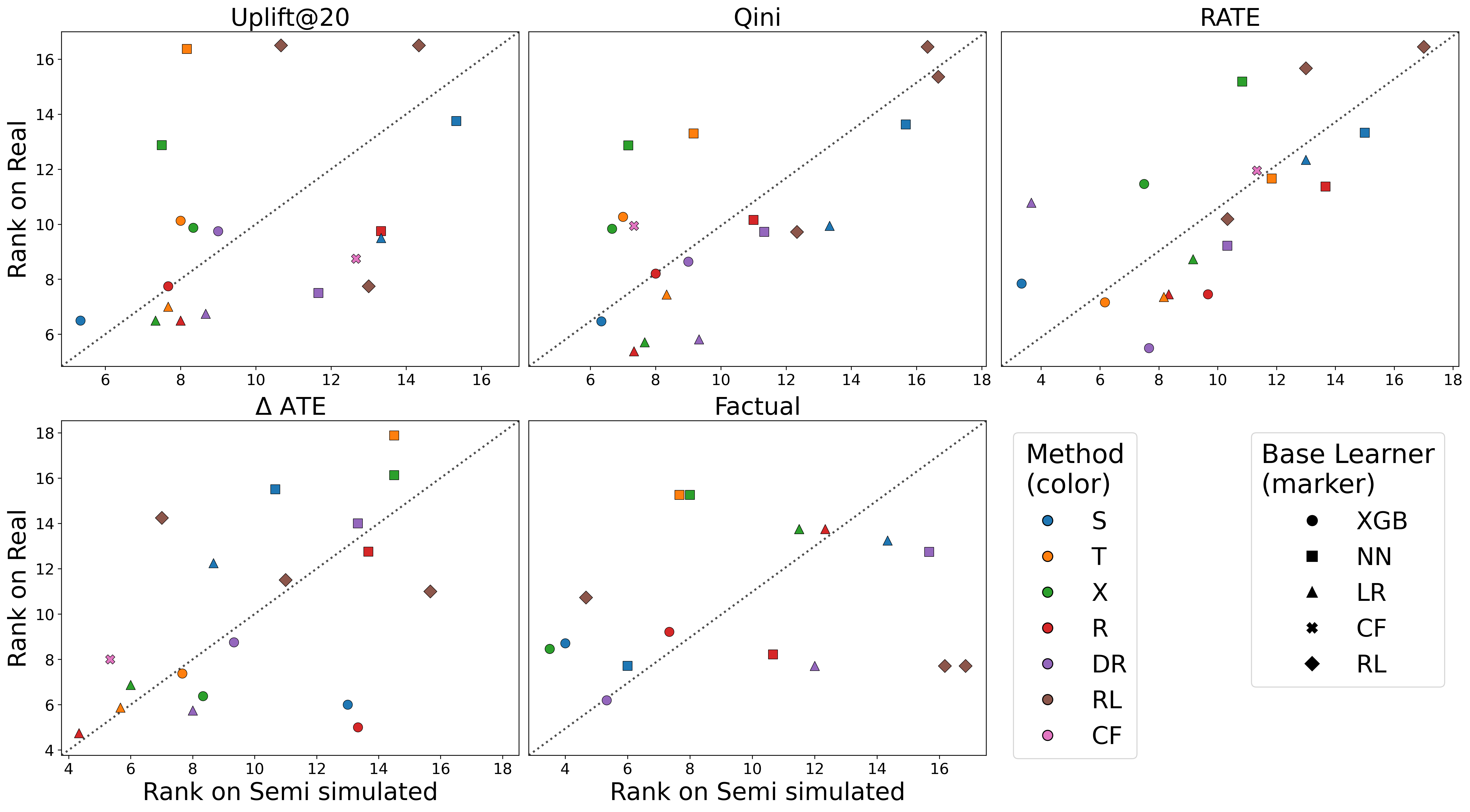}
    \caption{Observable-metric rankings on semi-simulated (x-axis) vs real (y-axis) datasets. Each point corresponds to one method. RL stands for representation learning and includes TARNet, CFR and DragonNet.}
    \label{fig:semi_real_transfer_scatter}
\end{figure}

Table~\ref{tab:real_top_methods}, together with Table~\ref{tab:simulated_top_methods_all}, suggests that semi-simulated and real datasets induce different selection patterns even when evaluated with observable metrics. We see for example that prevalent methods on semi-simulated benchmark, such as DragonNet and DR(LR) on IHDP, and R(LR) on Twins, do not appear at all in the top of real data. On a side note, XGBoost-based learners appear most prominently on ACIC2016 and NHEFS, two datasets of similar scale. We therefore can deduce that some differences do not stem from the type of the dataset but also, from their inherent characteristics. To quantify the difference more directly, Figure~\ref{fig:semi_real_transfer_scatter} compares the rank of each method on semi-simulated datasets against its rank on real datasets, using only observable metrics. If semi-simulated benchmarks were reliable proxies for real evaluation, the points would concentrate close to the diagonal. Instead, the figure shows substantial dispersion: methods with similar standing on semi-simulated data can have different standing on real data, and vice versa. 

\begin{figure}[t]
    \centering
    \includegraphics[width=\linewidth]{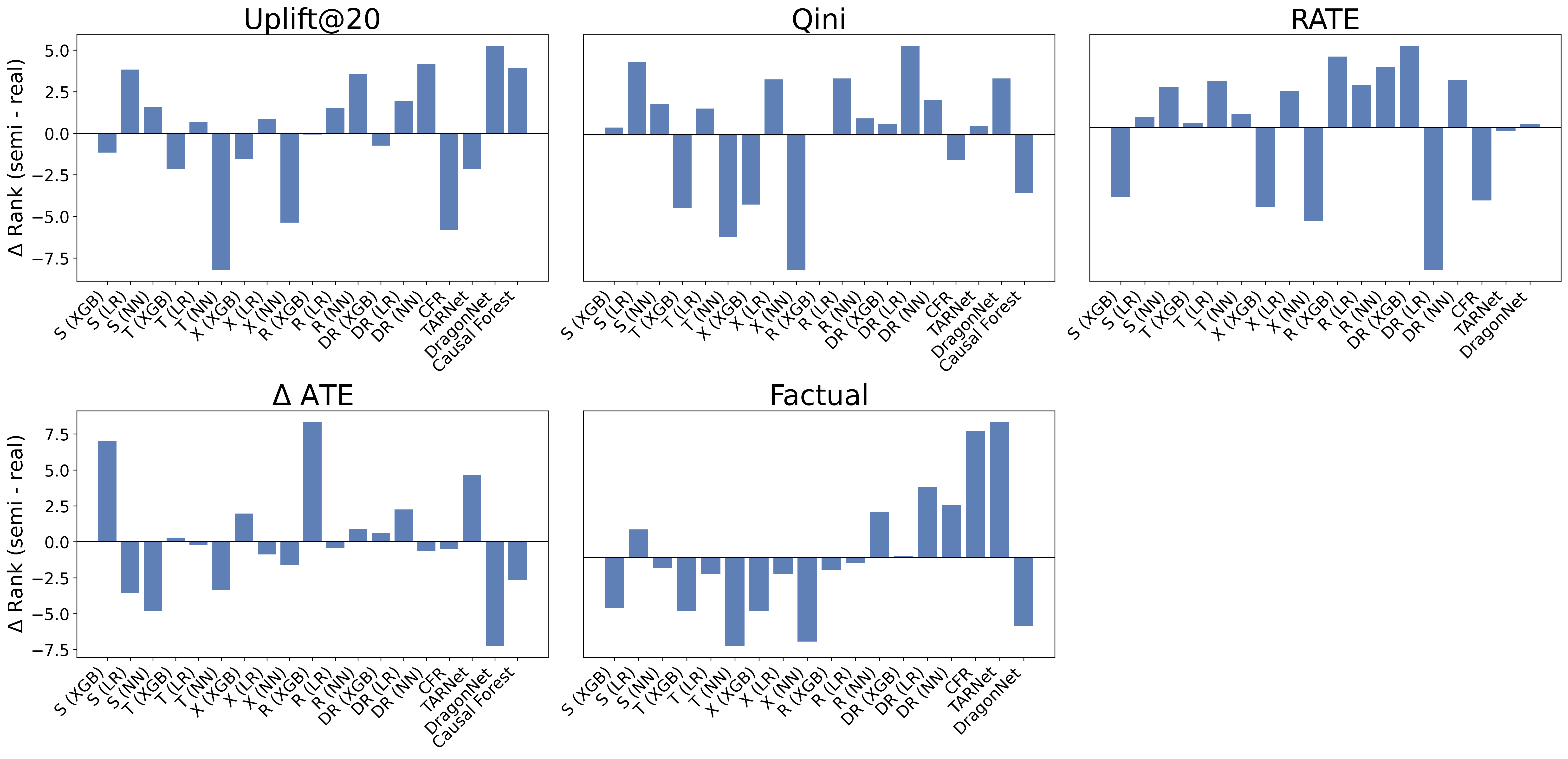}
    \caption{Average rank shift $\Delta \mathrm{Rank}=\mathrm{Rank}_{\text{semi}}-\mathrm{Rank}_{\text{real}}$ from semi-simulated to real datasets for each observable metric. Positive indicates rank improvement from semi-simulated to real, and vice versa.}
    \label{fig:delta_rank_method_base_per_metric}
\end{figure}

\begin{figure}[t]
    \centering
    \includegraphics[width=1\linewidth]{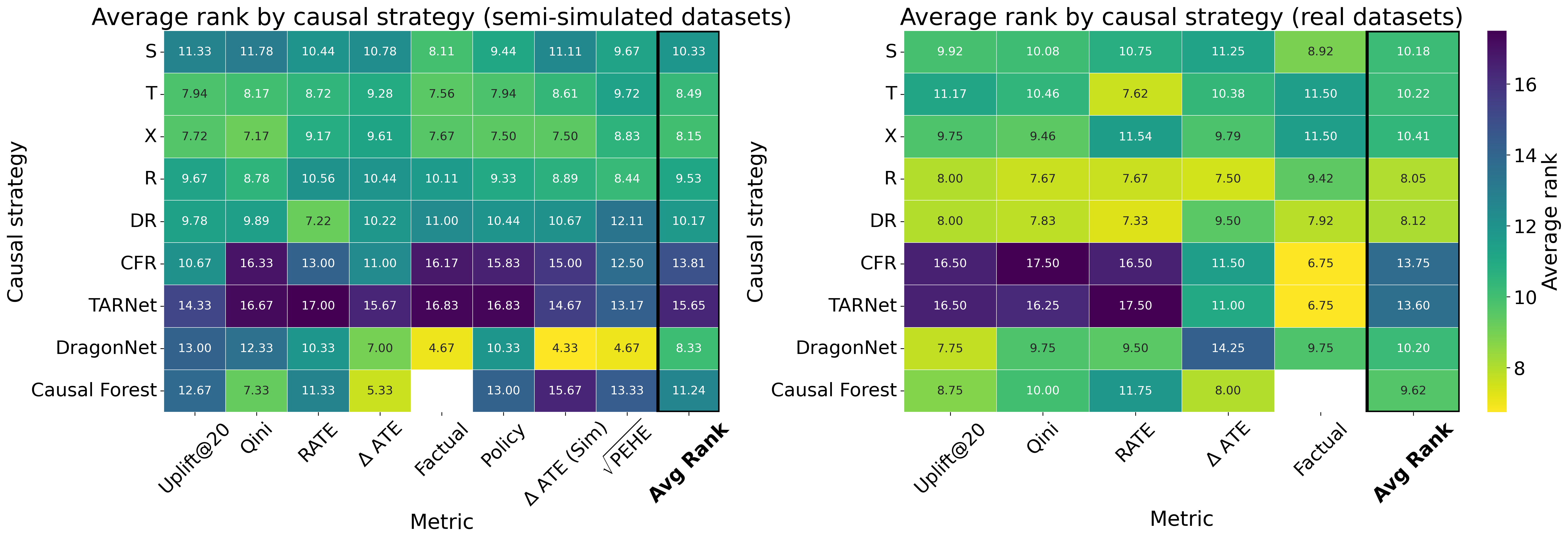}
    \caption{Average rank by causal strategy on semi-simulated and real datasets. Lower is better. }
    \label{fig:avg_rank_strategy_semi_real}
\end{figure}

Figure~\ref{fig:delta_rank_method_base_per_metric} provides a complementary view through the rank difference $\Delta \mathrm{Rank}=\mathrm{Rank}_{\text{semi}}-\mathrm{Rank}_{\text{real}}$. We see some substantial shifts, for example several R- and DR-based estimators improve under real-data ranking, whereas several T- and X-learners deteriorate. The rest of the methods undergo mostly metric-dependent shifts, indicating an overall significant and unstructured disagreement.  
Finally, Figure~\ref{fig:avg_rank_strategy_semi_real} shows that the preferred causal strategy also changes across benchmark families. On semi-simulated datasets, the best average strategies are the X-learner and DragonNet, followed by the T-learner, while in real datasets, the DR-learner and R-learner achieve the best average ranks. 

Taken together, these results show that conclusions drawn from semi-simulated benchmarks do not reliably transfer to real datasets, even when the comparison is restricted to observable metrics. Semi-simulated benchmarks can therefore be useful stress tests, but should not be treated as sufficient guides for practical applications. Their data-generating processes can impose ranking patterns that favor estimators whose inductive biases align with the simulation mechanism rather than with real treatment assignment or outcome noise. Consequently, the community faces not only a metric problem, but also a dataset problem.

\begin{figure}[h!]
    \centering
    \includegraphics[width=.9\linewidth]{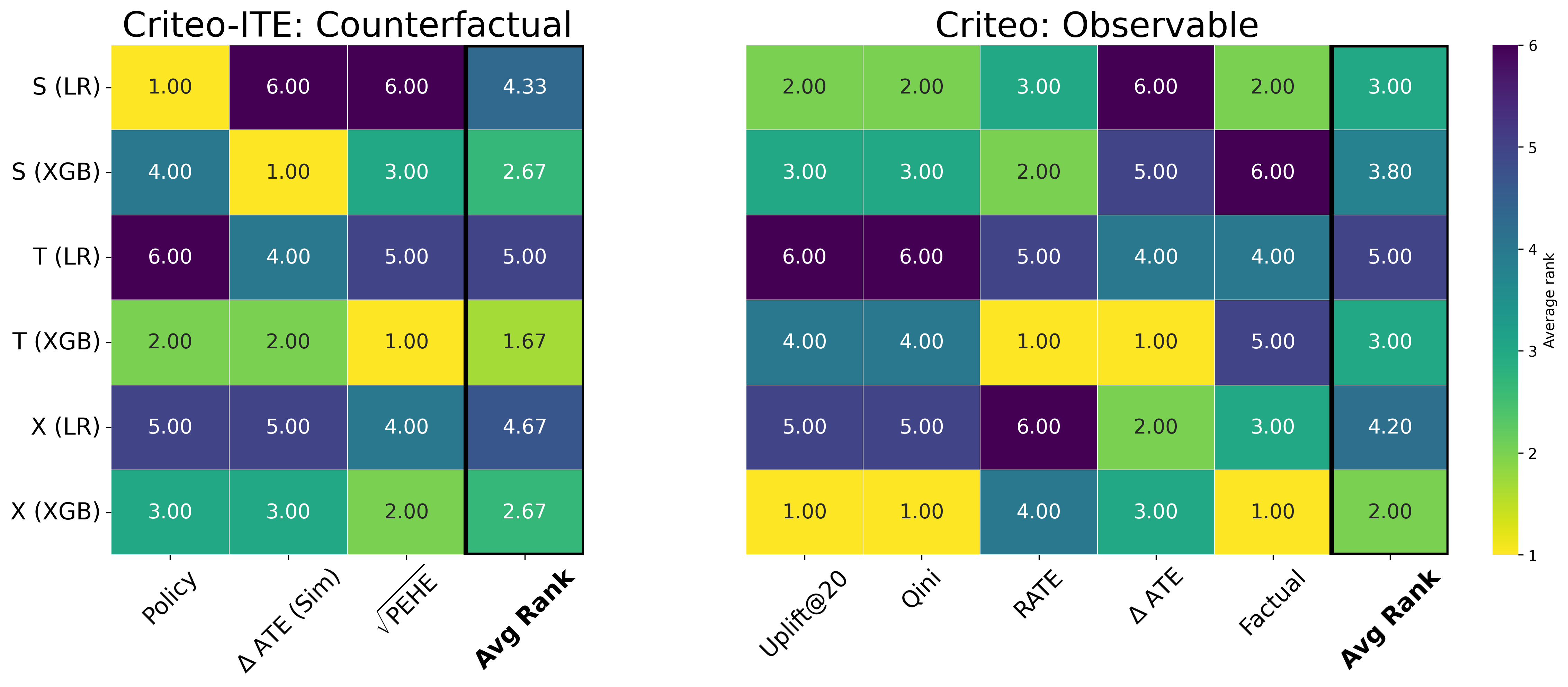}
    \caption{Matched Criteo case study. Left: average ranks of S-, T-, and X-learners with XGBoost and linear base learners under counterfactual metrics on Criteo-ITE. Right: average ranks of the same estimators under observable metrics on real Criteo. Lower rank is better.}
\label{fig:criteo_case_study}
\end{figure}

\subsection{Industrial-scale matched case study}
\label{sec:results_criteo_case_study}
We consider the real Criteo conversion dataset~\cite{diemert2018large} and the semi-simulated Criteo-ITE derived from the same industrial data source~\cite{diemert2021large}. 
The two benchmarks share the same covariates and treatment assignments, while the outcome-generating process differs: Criteo uses observed conversion outcomes, whereas Criteo-ITE replaces the outcome with synthetic potential outcomes generated from the same feature matrix. 
This provides a large-scale matched case study that controls for the covariate space and application domain. Criteo-ITE with counterfactual-based metrics represents the evaluation protocol of most academic research papers, while Criteo with observable metrics depicts the industrial setting.
Due to computational constraints, we restrict the comparison to S-, T-, and X-learners with XGBoost and linear base learners and do not repeat it for multiple seeds.

Figure \ref{fig:criteo_case_study} shows that the average ranking changes between the two settings. Under counterfactual metrics on Criteo-ITE, the best average rank is obtained by T(XGB), followed by S(XGB) and X(XGB), which are tied. On real Criteo with observable metrics, the best average rank is obtained by X(XGB), followed by S(LR) and T(XGB), which are tied. 
Thus, the matched Criteo setting shows partial transfer: several methods favored by the research protocol remain competitive on the real setting, especially XGBoost-based meta-learners. However, the ordering still changes, and the best method under counterfactual semi-simulated evaluation is not the best method under observable real-data evaluation. This partial overlap is also expected because this case study uses a much smaller estimator set than the main benchmark. This should therefore be interpreted as incomplete, rather than absent, transfer: even in this favorable matched setting, with shared covariates and treatment assignments, counterfactual evaluation is not a fully robust proxy for observable real-data evaluation.

\subsection{Practical takeaways}
\label{sec:results_practical_takeaways}

\begin{table}[h!]
\centering
\small
\begin{tabular}{lccc}
\toprule
\textbf{Regime} & \textbf{Best} & \textbf{Second} & \textbf{Third} \\
\midrule
Counterfactual metrics on semi-simulated benchmarks & X (XGB) & T (XGB) & DragonNet \\
Observable metrics on real datasets & S (XGB) & R (LR)/DR (LR) & R (XGB) \\
Overall & T (XGB) & X (XGB) & S (XGB) \\
\bottomrule
\end{tabular}
\caption{Top-ranked methods by evaluation regime. }
\label{tab:top_by_regime}
\end{table}

The previous results show that there is no universally best estimator across metrics and datasets. Nevertheless, some practical patterns emerge. First, simple XGBoost-based meta-learners are consistently strong, as indicated in Table~\ref{tab:top_by_regime} and in Appendix \ref{app:rank_heatmaps}.  T(XGB), X(XGB), and S(XGB) are the three best methods in the overall ranking.
T(XGB) and X(XGB) are strongly favored by counterfactual semi-simulated metrics, whereas S(XGB) is the most stable method across regimes and is the best-ranked method on real data. 
Second, the detailed average-rank heatmaps in Appendix~\ref{app:rank_heatmaps}
show that specialized neural architectures underperform overall. DragonNet is the most competitive among them, but solely under counterfactual semi-simulated evaluation. 

Hence, methodological work that aims to improve CATE estimation should include simple meta-learners with strong predictive backbones as benchmarks. Moreover, the empirical gains from new causal strategies have to be carefully controlled to remove the role of the base learner, which is very strong.
Finally, the counterfactual metrics and semi-simulated data regime contrasts sharply with the observable and real data regime since they have no method in common in Table~\ref{tab:top_by_regime}.
This supports our overall conclusion that the methods favored by the dominant evaluation protocol in current causal ML research can differ from those favored under observable, deployment-oriented evaluation.

\section{Conclusion}
\label{sec:conclusion}

We studied the alignment between the evaluation regimes commonly used in causal ML research and those available in real-world applications. Across 117 semi-simulated and real benchmark instances, our results show that counterfactual metrics on semi-simulated data do not reliably select the estimators preferred by observable ranking and outcome metrics. We further find that conclusions drawn from semi-simulated benchmarks do not reliably transfer to real datasets, even when the comparison is restricted to observable metrics. These findings suggest that the field faces both a metric problem and a dataset problem: oracle-style evaluation can be useful for controlled analysis, but it should not be treated as a sufficient proxy for deployment-oriented validation. Our results also reinforce the importance of strong, simple baselines such as meta-learners with competitive tabular predictive models. Future work should develop evaluation metrics and benchmark protocols that better connect oracle-style counterfactual evaluation with observable deployment-oriented criteria. Extending this analysis to continuous and multi-valued treatments is another important direction, since evaluation in these settings is even less standardized than in the binary-treatment case.

\paragraph{Limitations.}
Our study is empirical and, therefore, necessarily limited by the datasets, estimators, and metrics considered. We focus on binary treatment-effect estimation and primarily tabular benchmarks. Real datasets allow only observable evaluation, so they cannot reveal whether a method truly recovers individual treatment effects. Some comparisons are also constrained by implementation and computational considerations: for example, the industrial-scale matched Criteo case study includes a smaller estimator set and fewer repetitions than the other benchmarks. 
Finally, while our results identify systematic misalignment between common evaluation regimes, they do not prescribe a single universally best metric or estimator. 

\paragraph{Code.}
Code to reproduce the benchmark experiments available at
\url{https://github.com/geopanag/real_vs_semisim}.

\bibliographystyle{plainnat}
\bibliography{bib}

\appendix

\section*{Appendix}
\addcontentsline{toc}{section}{Appendix}

\section{Metric definitions}
\label{app:metric_definitions}

Let $\hat{\tau}(x)$ denote the predicted treatment effect. For ranking-based
metrics, let $S(x)$ denote the prioritization score induced by the model, where
larger values indicate higher predicted treatment benefit. In our experiments we
use $S(x)=\hat{\tau}(x)$. Let $F_S$ be the cumulative distribution function of
$S(X)$. Then $F_S(S(X_i))$ denotes the quantile rank of unit $i$ under the
prioritization rule, and units satisfying $F_S(S(X_i)) \geq 1-u$ correspond to
the top $u$ fraction of the population.

\subsection{Ranking-based metrics}

Ranking metrics evaluate whether a model correctly prioritizes units for
treatment. Let $S(X_i)$ denote the prioritization score for test unit $i$, where
larger values indicate larger predicted treatment benefit. We sort the test units
in decreasing order of $S(X_i)$ and denote the resulting ordering by
$i_{(1)},\ldots,i_{(n)}$. For a treatment fraction $u\in(0,1]$, let
\[
A_u(S)
=
\left\{
i_{(1)},\ldots,i_{(\lceil un\rceil)}
\right\}
\]
be the top-$u$ fraction of test units according to the model ranking.

For any subset $A$ of test units, we compute the empirical treatment-control
difference in observed outcomes as
\[
\widehat{\tau}(A)
=
\frac{1}{n_1(A)}
\sum_{i\in A:T_i=1} Y_i
-
\frac{1}{n_0(A)}
\sum_{i\in A:T_i=0} Y_i,
\]
where
\[
n_1(A)=\sum_{i\in A}\mathbf{1}\{T_i=1\},
\qquad
n_0(A)=\sum_{i\in A}\mathbf{1}\{T_i=0\}.
\]
Thus, these metrics use only observed test-set outcomes and treatment
assignments. They do not require observing individual counterfactual outcomes.

\paragraph{Up@20.}
Up@20 \citep{rafla2022evaluation} computes the empirical treatment-control
difference among the top 20\% of test units ranked by the predicted treatment
effect:
\[
\widehat{\mathrm{Up@20}}(S)
=
\widehat{\tau}\!\left(A_{0.2}(S)\right).
\]

\paragraph{Qini score.}
The Qini score \citep{diemert2018large} measures the cumulative gain from
targeting units according to the model ranking, relative to random targeting.
For a treatment fraction $u$, the empirical cumulative gain is
\[
\widehat{G}(u;S)
=
\frac{|A_u(S)|}{n}
\widehat{\tau}\!\left(A_u(S)\right),
\]
whereas the gain under random targeting at the same treatment fraction is
\[
\widehat{G}_{\mathrm{rand}}(u)
=
\frac{|A_u(S)|}{n}
\widehat{\tau}\!\left(\{1,\ldots,n\}\right).
\]
The empirical Qini curve is therefore
\[
\widehat{Q}(u;S)
=
\widehat{G}(u;S)
-
\widehat{G}_{\mathrm{rand}}(u),
\]
and the Qini score is the area under this curve:
\[
\widehat{\mathrm{Qini}}(S)
=
\int_0^1
\widehat{Q}(u;S)\,du.
\]
In practice, the integral is approximated over the empirical ranking.

\paragraph{RATE AUTOC.}
RATE AUTOC \citep{yadlowsky2025evaluating} is based on the targeting operator
characteristic (TOC) curve. For a prioritization rule $S$, the empirical TOC at
treatment fraction $u$ is
\[
\widehat{\mathrm{TOC}}(u;S)
=
\widehat{\tau}\!\left(A_u(S)\right)
-
\widehat{\tau}\!\left(\{1,\ldots,n\}\right).
\]
It compares the treatment-control difference among the top-$u$ ranked units to
the treatment-control difference in the full test set. The AUTOC score is the
area under this curve:
\[
\widehat{\mathrm{AUTOC}}(S)
=
\int_0^1
\widehat{\mathrm{TOC}}(u;S)\,du.
\]
In practice, the integral is approximated over the empirical ranking. Compared
with Qini, AUTOC places relatively more emphasis on the highest-ranked units,
whereas Qini weights the TOC by the treated fraction.

\subsection{Outcome-based metrics}

\paragraph{Factual loss.}
Factual loss \citep{curth2023search} measures prediction error on the observed
outcome under the factual treatment assignment:
\[
\mathcal{L}_{\mathrm{factual}}
=
\sum_i
\left(
Y_i-\hat{Y}_i(T_i)
\right)^2 .
\]

\paragraph{Difference in ATE.}
$\Delta\mathrm{ATE}$ \citep{curth2021really} measures the squared discrepancy
between the estimated and empirical average treatment effect:
\[
\Delta\mathrm{ATE}
=
\left(
\tau-\hat{\tau}
\right)^2 .
\]

\subsection{Counterfactual-based metrics}

\paragraph{PEHE.}
Precision in estimation of heterogeneous effects (PEHE)
\citep{alaa2017bayesian} measures individual-level treatment-effect error. We
report its square root:
\[
\sqrt{\mathrm{PEHE}}
=
\sqrt{
\frac{1}{n}
\sum_i
\left(
\hat{\tau}(x_i)-\tau(x_i)
\right)^2
}.
\]

\paragraph{Policy value/risk.}
Policy value/risk \citep{curth2021really}, also referred to as decision-support
evaluation, evaluates the expected outcome under the treatment policy induced by
the estimated treatment effects:
\[
\mathbb{E}
\left[
\pi(x)Y(1)
+
(1-\pi(x))Y(0)
\right],
\]
where
\[
\pi(x)=\mathbb{I}\left(\hat{\tau}(x)>0\right).
\]

\paragraph{Difference in simulated ATE.}
$\Delta\mathrm{ATE}(\mathrm{Sim})$ \citep{johansson2016learning} measures the
squared discrepancy between the estimated ATE and the true ATE using simulated
potential outcomes:
\[
\Delta\mathrm{ATE}(\mathrm{Sim})
=
\left(
\mathbb{E}[\hat{\tau}(x)]
-
\mathbb{E}[\tau(x)]
\right)^2 .
\]

\section{Methods and Hyperparameter ranges}
\label{app:hyperparameters}

We evaluate a diverse set of causal effect estimation approaches, including meta-learners, representation learning methods, and tree-based models.

\paragraph{Meta-learners.}
We consider standard meta-learning strategies for conditional average treatment effect (CATE) estimation \citep{kunzel2019metalearners}. 
\begin{itemize}
    \item The \textsc{S-learner} fits a single model to predict outcomes given covariates and treatment. 
\item The \textsc{T-learner} fits separate models for the treated and control groups and computes the difference in predictions. 
\item The \textsc{X-learner} improves upon the T-learner by leveraging imputed treatment effects and reweighting based on propensity scores. 
\item The \textsc{R-learner} formulates CATE estimation as a residual-on-residual regression problem, allowing for flexible nuisance estimation. 
\item The doubly robust (\textsc{DR}) combines outcome and propensity models to provide consistent estimates when at least one of the nuisance components is correctly specified.
\end{itemize}
We pair each metalearning methodology with different base learners to study the impact of various inductive biases:

\begin{itemize}
 \item \textsc{XGB}: gradient boosted decision trees.
    \item \textsc{NN}: a two-layer multilayer perceptron (MLP) with ReLU activations.
    \item \textsc{LR}: linear (or logistic) regression depending on the task.
\end{itemize}

\paragraph{Non meta-learners}
We additionally consider a set of widely used tailored benchmarks. 

\begin{itemize}
\item \textsc{TARNet} \citep{shalit2017estimating} learns a shared representation of covariates with separate outcome heads for treated and control units. 
\item \textsc{CFR} (Counterfactual Regression) \citep{shalit2017estimating} extends TARNet by adding a distribution matching regularization to balance treated and control representations, removing potential selection bias. 
\item \textsc{DragonNet} \citep{shi2019adapting} further incorporates propensity estimation into the architecture to improve treatment effect estimation via targeted regularization.
 \item \textsc{CausalForest} \citep{wager2018estimation} is a tree-based method that adapts random forests to CATE estimation by splitting the leaves such that the treatment effect differences is maximized.
\end{itemize}

\paragraph{Hyperparameter Ranges}
Hyperparameters are selected by inner cross-validation on the training split only. 
For classification outcomes, the inner objective is negative log-loss; for regression outcomes, the corresponding regression loss is used. 
The hyperparameter ranges are reported in Table~\ref{tab:hyperparameter_ranges}.

\begin{table}[t]
\centering
\small
\setlength{\tabcolsep}{4pt}
\begin{tabular}{llll}
\toprule
\textbf{Base learner family} & \textbf{Task} & \textbf{Hyperparameter} & \textbf{Candidate values} \\
\midrule

\multirow{2}{*}{Linear model}
& Classification 
& Logistic regularization strength $C$ 
& \makecell[l]{$\{10^{-4},10^{-3},10^{-2},$\\$10^{-1},1,10\}$} \\

& Regression 
& Ridge penalty $\alpha$ 
& \makecell[l]{$\{10^{-4},10^{-3},10^{-2},$\\$10^{-1},1,10,10^{2},10^{3},10^{4}\}$} \\

\midrule

\multirow{4}{*}{XGBoost}
& Classification / Regression 
& Number of trees 
& $\{100,300\}$ \\

& Classification / Regression 
& Maximum depth 
& $\{3,6\}$ \\

& Classification / Regression 
& Learning rate 
& $\{0.05,0.1\}$ \\

& Classification / Regression 
& Subsample ratio 
& $\{0.8,1.0\}$ \\

\midrule

\multirow{4}{*}{Neural network}
& Classification / Regression 
& Hidden layers 
& $\{(64,32),(128,64)\}$ \\

& Classification / Regression 
& $\ell_2$ penalty $\alpha$ 
& $\{10^{-4},10^{-3},10^{-2}\}$ \\

& Classification 
& Learning rate 
& $\{10^{-3},5\cdot 10^{-3}\}$ \\

& Regression 
& Learning rate 
& $\{10^{-3},5\cdot 10^{-4}\}$ \\

\bottomrule
\end{tabular}
\caption{Hyperparameter ranges used for the predictive base learners inside the meta-learners. Hyperparameters are selected by inner cross-validation on the training split. Logistic classification uses 6 candidate values for $C$, ridge regression uses 9 candidate values for $\alpha$, XGBoost uses $2\times2\times2\times2=16$ configurations, and neural networks use $2\times3\times2=12$ configurations.}
\label{tab:hyperparameter_ranges}
\end{table}

Combining these grids with the five meta-learners yields
\[
5 \times (6 + 16 + 12) = 170
\]
configurations for classification outcomes and
\[
5 \times (9 + 16 + 12) = 185
\]
configurations for regression outcomes.
meta-learner/base-learner/hyperparameter configurations. 
The four non-meta-learner baselines (CFR, TARNet, DragonNet, and Causal Forest) are used with their default parameters.

\section{Statistical test for ranking--counterfactual metric agreement}
\label{app:wilcoxon}

To test whether deployment-oriented ranking metrics agree more with one another than with counterfactual metrics, we compare metric-induced method rankings rather than raw metric values, since the metrics have different scales and directions. Let
\[
\mathcal{M}_{\mathrm{rank}}
=
\{\textsc{Uplift@20}, \mathrm{Qini}, \mathrm{RATE}\}
\]
denote the ranking metrics, and let
\[
\mathcal{M}_{\mathrm{cf}}
=
\{\sqrt{\mathrm{PEHE}}, \mathrm{Policy}, \Delta\mathrm{ATE}(\mathrm{Sim})\}
\]
denote the counterfactual metrics. For each semi-simulated benchmark instance $b$, we first average each method's score over random seeds and then rank all methods separately under each metric.

We compute Kendall's $\tau$ between all pairs of metric-induced rankings. Let $\mathcal{P}_{\mathrm{rank,rank}}$ be the set of ranking--ranking metric pairs and $\mathcal{P}_{\mathrm{rank,cf}}$ the set of ranking--counterfactual metric pairs. We define
\[
\bar{\tau}^{(b)}_{\mathrm{rank,rank}}
=
\frac{1}{|\mathcal{P}_{\mathrm{rank,rank}}|}
\sum_{(m_i,m_j)\in \mathcal{P}_{\mathrm{rank,rank}}}
\tau_b(m_i,m_j),
\]
and
\[
\bar{\tau}^{(b)}_{\mathrm{rank,cf}}
=
\frac{1}{|\mathcal{P}_{\mathrm{rank,cf}}|}
\sum_{(m_i,m_j)\in \mathcal{P}_{\mathrm{rank,cf}}}
\tau_b(m_i,m_j).
\]
The paired difference for benchmark instance $b$ is
\[
D_b
=
\bar{\tau}^{(b)}_{\mathrm{rank,rank}}
-
\bar{\tau}^{(b)}_{\mathrm{rank,cf}}.
\]
Positive values of $D_b$ indicate that ranking metrics agree more with one another than they agree with counterfactual metrics. We test
\[
H_0: \mathrm{median}(D_b)=0
\]
against the one-sided alternative
\[
H_1: \mathrm{median}(D_b)>0
\]
using a Wilcoxon signed-rank test across benchmark instances $b$.

\section{Detailed average-rank heatmaps}
\label{app:rank_heatmaps}

\begin{figure}[h!]
    \centering
    \includegraphics[width=\linewidth]{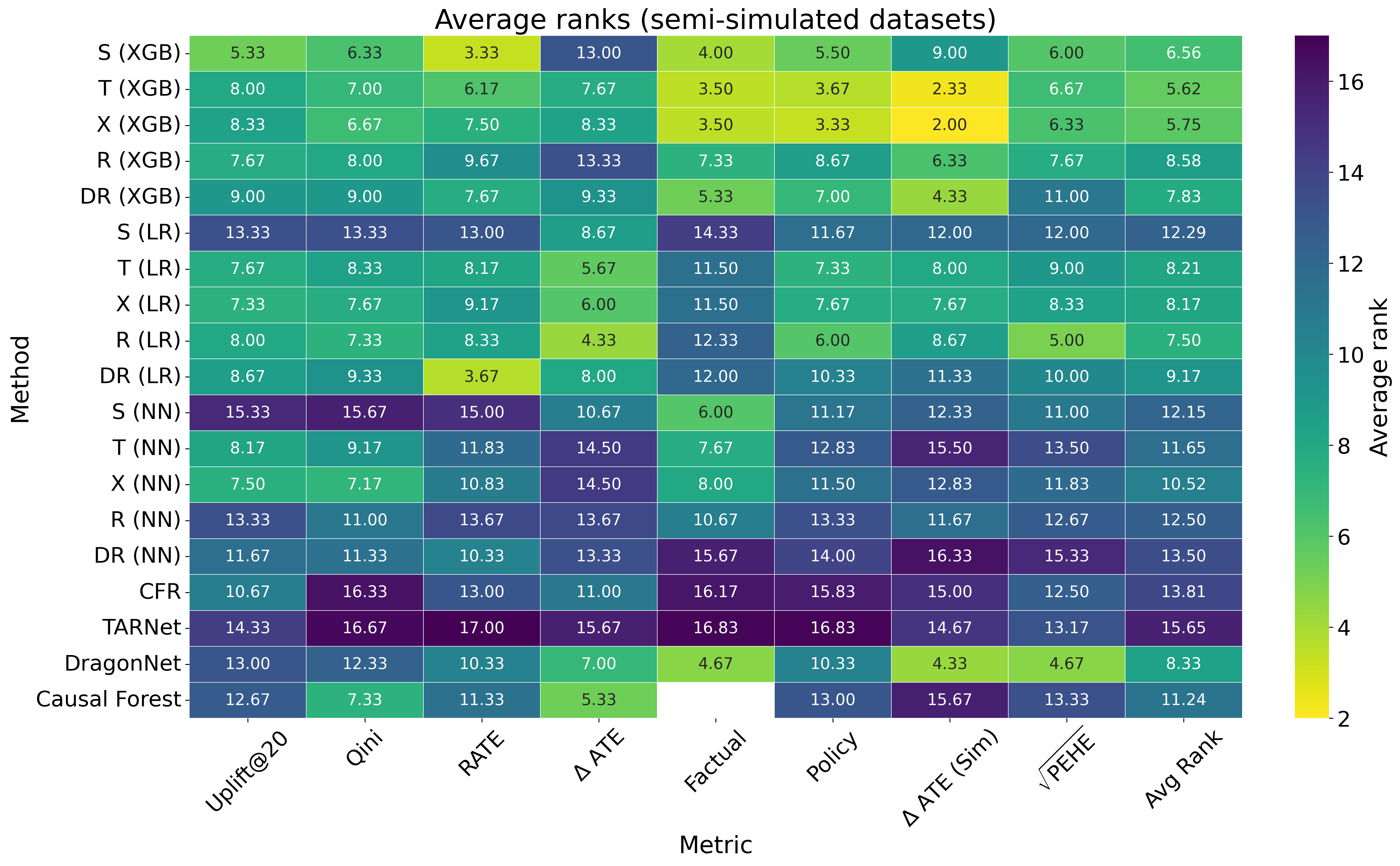}
    \caption{Average ranks of all methods on the semi-simulated datasets across all evaluation metrics. Lower ranks are better. The final column reports the average rank across metrics.}
    \label{fig:rank_heatmap_semi_appendix}
\end{figure}

\begin{figure}[h!]
    \centering
    \includegraphics[width=\linewidth]{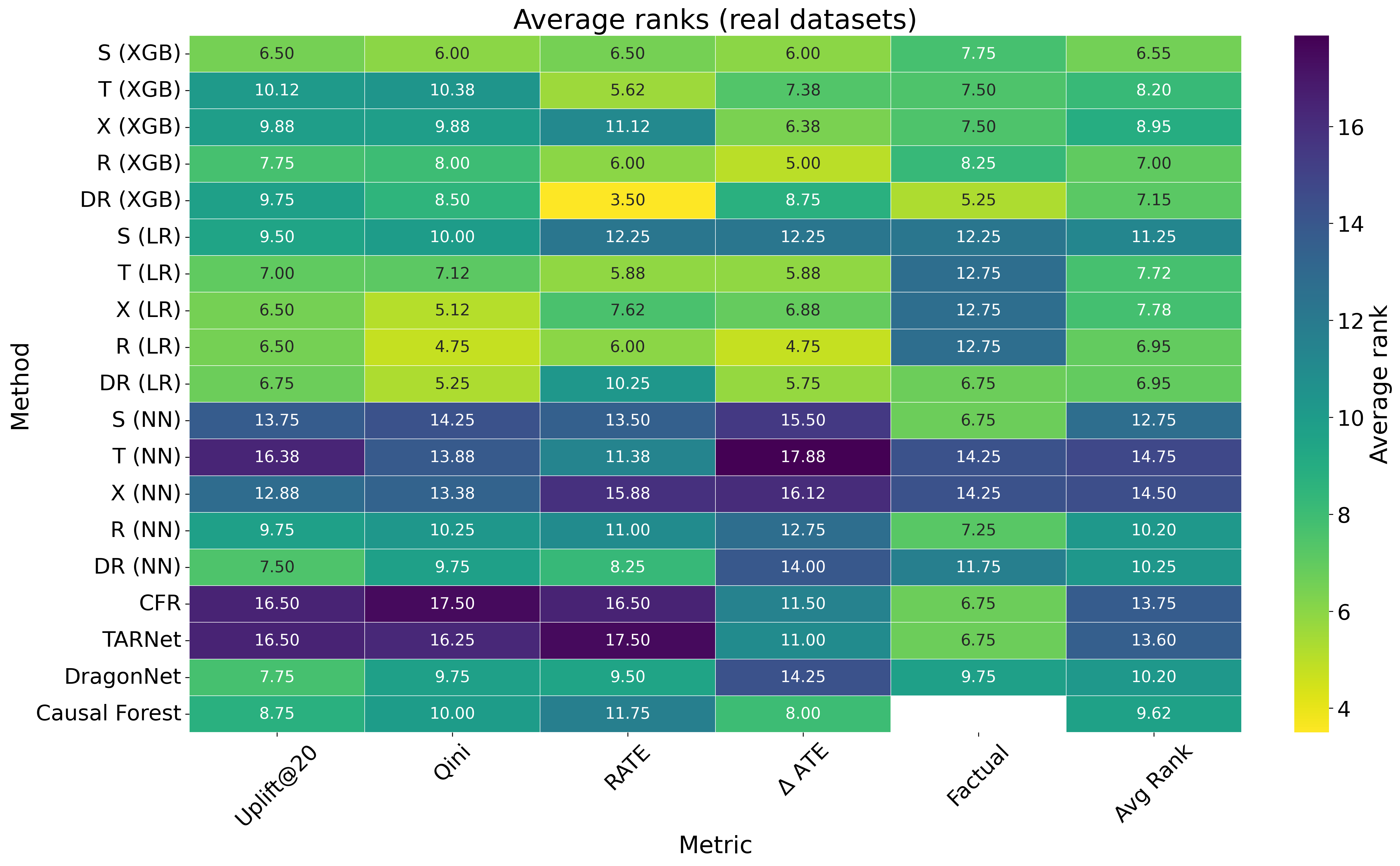}
    \caption{Average ranks of all methods on the real datasets across observable evaluation metrics. Lower ranks are better. The final column reports the average rank across metrics.}
    \label{fig:rank_heatmap_real_appendix}
\end{figure}

To complement the practical takeaways in Section~\ref{sec:results_practical_takeaways}, Figures~\ref{fig:rank_heatmap_semi_appendix} and~\ref{fig:rank_heatmap_real_appendix} report the full average-rank heatmaps for semi-simulated and real datasets, respectively. Each entry shows the average rank of a method under a given metric, where lower is better. The last column reports the average rank across all metrics available in the corresponding benchmark family.

The semi-simulated heatmap in Figure~\ref{fig:rank_heatmap_semi_appendix} shows that simple meta-learners with XGBoost backbones dominate the overall ranking, with T(XGB), X(XGB), and S(XGB) obtaining the strongest average performance. It also highlights the regime-specific strengths of some methods: for example, DragonNet is relatively competitive on counterfactual and outcome-style metrics, but much weaker on ranking metrics. Neural-network base learners and specialized neural architectures such as CFR and TARNet tend to have substantially weaker average ranks.

The real-data heatmap in Figure~\ref{fig:rank_heatmap_real_appendix} shows a different pattern. While S(XGB) remains the strongest overall method, several R- and DR-based estimators become more competitive, especially R(LR), DR(LR), R(XGB), and DR(XGB). The figure also confirms that methods with neural-network backbones remain generally weak on these tabular benchmarks. Overall, the detailed heatmaps support the practical conclusions drawn in the main text: strong tabular predictive backbones are critical, and the methods favored by semi-simulated counterfactual evaluation are not the same as those that perform best under observable real-data evaluation.

\section{Computational resources.}
\label{app:computational_resources}
All experiments were 
run on one CPU node with 512 GB RAM and 0.5 TB storage. The time of execution varied. Most experiments take less than a day, with the exception of Criteo datasets where the XGB-X combination took up to 5 days to run, and the simpler methods took 3 days.

\section{Detailed metric results}
\label{app:detailed_results}

\begin{table}[h!]
\centering
\begin{tabular}{lrrrrrrrr}
\toprule
Method & Uplift@20 & Qini & RATE & $\Delta$ ATE & Factual & Policy & $\Delta$ ATE (Sim) & $\sqrt{\mathrm{PEHE}}$ \\
\midrule
CFR & 4.76 & 0.50 & -0.22 & 0.80 & 3.63 & 3.06 & 0.90 & 4.42 \\
Causal Forest & 8.20 & 0.63 & 0.21 & 0.65 & NaN & 3.13 & 1.45 & 3.68 \\
DR (LR) & 7.41 & 0.61 & 2.93 & 0.83 & 2.16 & 3.48 & 0.59 & 3.49 \\
DR (NN) & 7.41 & 0.60 & 1.95 & 0.87 & 2.22 & 3.54 & 0.52 & 3.04 \\
DR (XGB) & 8.68 & 0.63 & 3.19 & 0.94 & 0.98 & 3.76 & 0.25 & 3.68 \\
DragonNet & 8.43 & 0.63 & 2.42 & 0.96 & 1.20 & 3.75 & 0.15 & 1.87 \\
R (LR) & 7.69 & 0.61 & 2.01 & 0.74 & 2.17 & 3.62 & 0.38 & 2.53 \\
R (NN) & 7.77 & 0.62 & 1.78 & 0.92 & 2.02 & 3.60 & 0.33 & 2.79 \\
R (XGB) & 8.58 & 0.64 & 2.48 & 1.26 & 1.06 & 3.77 & 0.26 & 1.64 \\
S (LR) & 4.66 & 0.49 & 0.10 & 0.77 & 2.32 & 3.45 & 0.33 & 3.39 \\
S (NN) & 6.38 & 0.57 & 1.26 & 0.89 & 1.97 & 3.50 & 0.36 & 3.07 \\
S (XGB) & 8.57 & 0.63 & 2.59 & 1.04 & 0.95 & 3.77 & 0.14 & 1.42 \\
T (LR) & 7.42 & 0.61 & 2.05 & 0.71 & 2.16 & 3.63 & 0.42 & 2.64 \\
T (NN) & 7.88 & 0.62 & 2.09 & 0.84 & 1.92 & 3.60 & 0.38 & 2.90 \\
T (XGB) & 8.87 & 0.63 & 2.59 & 0.96 & 0.93 & 3.78 & 0.13 & 1.29 \\
TARNet & 6.87 & 0.57 & -0.30 & 1.07 & 2.64 & 2.99 & 0.51 & 4.08 \\
X (LR) & 7.42 & 0.61 & 2.02 & 0.71 & 2.16 & 3.63 & 0.41 & 2.59 \\
X (NN) & 7.88 & 0.62 & 2.04 & 0.90 & 1.92 & 3.60 & 0.34 & 2.63 \\
X (XGB) & 8.71 & 0.63 & 2.53 & 1.01 & 0.93 & 3.78 & 0.09 & 1.13 \\
\bottomrule
\end{tabular}
\caption{ACIC2016}
\label{tab:sim_acic2016_avg_metric_values}
\end{table}

\begin{table}[h!]
\centering
\begin{tabular}{lrrrrrrrr}
\toprule
Method & Uplift@20 & Qini & RATE & $\Delta$ ATE & Factual & Policy & $\Delta$ ATE (Sim) & $\sqrt{\mathrm{PEHE}}$ \\
\midrule
CFR & 9.16 & 0.57 & 3.11 & 0.69 & 1.74 & 17.04 & 0.45 & 2.84 \\
Causal Forest & 8.70 & 0.59 & 1.60 & 0.73 & NaN & 16.55 & 0.61 & 4.73 \\
DR (LR) & 10.16 & 0.65 & 3.79 & 0.58 & 1.66 & 17.11 & 0.25 & 2.24 \\
DR (NN) & 9.06 & 0.54 & 2.12 & 0.92 & 1.85 & 16.90 & 0.68 & 4.12 \\
DR (XGB) & 10.03 & 0.60 & 3.64 & 0.58 & 1.50 & 17.10 & 0.27 & 2.20 \\
DragonNet & 6.37 & 0.57 & 1.39 & 0.37 & 0.86 & 10.37 & 0.14 & 0.65 \\
R (LR) & 10.04 & 0.63 & 3.29 & 0.59 & 1.77 & 17.08 & 0.35 & 2.53 \\
R (NN) & 9.31 & 0.58 & 3.27 & 0.81 & 1.66 & 17.01 & 0.60 & 3.05 \\
R (XGB) & 9.74 & 0.61 & 2.41 & 0.70 & 1.75 & 17.04 & 0.40 & 2.63 \\
S (LR) & 3.09 & 0.47 & -0.58 & 1.12 & 2.17 & 15.90 & 0.80 & 5.73 \\
S (NN) & 8.83 & 0.59 & 3.20 & 0.98 & 1.58 & 16.86 & 0.68 & 3.91 \\
S (XGB) & 9.97 & 0.60 & 3.86 & 0.69 & 1.52 & 17.06 & 0.50 & 2.84 \\
T (LR) & 10.15 & 0.64 & 3.77 & 0.59 & 1.67 & 17.10 & 0.24 & 2.24 \\
T (NN) & 9.32 & 0.59 & 3.05 & 0.95 & 1.32 & 17.05 & 0.67 & 2.68 \\
T (XGB) & 10.02 & 0.62 & 3.73 & 0.54 & 1.33 & 17.14 & 0.21 & 1.74 \\
TARNet & 8.84 & 0.56 & 2.87 & 0.78 & 1.82 & 17.02 & 0.58 & 3.10 \\
X (LR) & 10.15 & 0.64 & 3.77 & 0.59 & 1.67 & 17.10 & 0.24 & 2.24 \\
X (NN) & 9.43 & 0.60 & 3.44 & 0.78 & 1.32 & 17.06 & 0.48 & 2.61 \\
X (XGB) & 10.02 & 0.62 & 3.53 & 0.54 & 1.33 & 17.14 & 0.21 & 1.74 \\
\bottomrule
\end{tabular}
\caption{IHDP}
\label{tab:sim_ihdp_avg_metric_values}
\end{table}

\begin{table}[h!]
\centering
\begin{tabular}{lrrrrrrrr}
\toprule
Method & Uplift@20 & Qini & RATE & $\Delta$ ATE & Factual & Policy & $\Delta$ ATE (Sim) & $\sqrt{\mathrm{PEHE}}$ \\
\midrule
CFR & -0.01 & -0.44 & -0.00 & 0.03 & 0.24 & 0.16 & 0.02 & 0.33 \\
Causal Forest & -0.03 & -0.36 & 0.01 & 0.01 & NaN & 0.17 & 0.01 & 0.33 \\
DR (LR) & -0.03 & -0.44 & -0.00 & 0.01 & 0.23 & 0.17 & 0.01 & 0.33 \\
DR (NN) & -0.02 & -0.29 & 0.01 & 0.02 & 0.20 & 0.17 & 0.02 & 0.33 \\
DR (XGB) & -0.05 & -0.43 & -0.01 & 0.01 & 0.17 & 0.17 & 0.01 & 0.33 \\
DragonNet & -0.04 & -0.48 & -0.00 & 0.01 & 0.17 & 0.17 & 0.01 & 0.33 \\
R (LR) & -0.03 & -0.39 & 0.00 & 0.01 & 0.18 & 0.18 & 0.01 & 0.32 \\
R (NN) & -0.04 & -0.43 & -0.01 & 0.02 & 0.19 & 0.17 & 0.02 & 0.33 \\
R (XGB) & -0.03 & -0.45 & -0.00 & 0.01 & 0.17 & 0.17 & 0.01 & 0.33 \\
S (LR) & -0.02 & -0.34 & 0.01 & 0.01 & 0.18 & 0.18 & 0.01 & 0.32 \\
S (NN) & -0.03 & -0.54 & -0.01 & 0.01 & 0.17 & 0.18 & 0.01 & 0.32 \\
S (XGB) & -0.02 & -0.35 & 0.00 & 0.02 & 0.17 & 0.18 & 0.01 & 0.32 \\
T (LR) & -0.03 & -0.42 & -0.00 & 0.01 & 0.18 & 0.17 & 0.01 & 0.33 \\
T (NN) & -0.02 & -0.37 & -0.00 & 0.04 & 0.19 & 0.17 & 0.03 & 0.33 \\
T (XGB) & -0.04 & -0.44 & -0.00 & 0.01 & 0.17 & 0.17 & 0.01 & 0.33 \\
TARNet & -0.03 & -0.46 & -0.01 & 0.03 & 0.24 & 0.16 & 0.02 & 0.33 \\
X (LR) & -0.03 & -0.42 & -0.00 & 0.01 & 0.18 & 0.17 & 0.01 & 0.33 \\
X (NN) & -0.02 & -0.37 & -0.00 & 0.04 & 0.19 & 0.17 & 0.03 & 0.33 \\
X (XGB) & -0.04 & -0.44 & -0.00 & 0.01 & 0.17 & 0.17 & 0.01 & 0.33 \\
\bottomrule
\end{tabular}
\caption{TWINS}
\label{tab:sim_twins_avg_metric_values}
\end{table}

\begin{table}[h!]
\centering
\begin{tabular}{lrrrrr}
\toprule
Method & Uplift@20 & Qini & RATE & $\Delta$ ATE & Factual \\
\midrule
CFR & 0.07 & 0.49 & 0.00 & 0.01 & 0.21 \\
Causal Forest & 0.08 & 0.51 & -0.00 & 0.01 & NaN \\
DR (LR) & 0.10 & 0.52 & 0.01 & 0.01 & 0.24 \\
DR (NN) & 0.08 & 0.50 & 0.02 & 0.01 & 0.24 \\
DR (XGB) & 0.09 & 0.51 & 0.01 & 0.01 & 0.24 \\
DragonNet & 0.08 & 0.52 & 0.00 & 0.01 & 0.24 \\
R (LR) & 0.10 & 0.51 & 0.01 & 0.01 & 0.24 \\
R (NN) & 0.09 & 0.50 & 0.01 & 0.02 & 0.23 \\
R (XGB) & 0.09 & 0.51 & 0.01 & 0.01 & 0.24 \\
S (LR) & 0.10 & 0.54 & 0.01 & 0.02 & 0.24 \\
S (NN) & 0.08 & 0.49 & -0.00 & 0.05 & 0.23 \\
S (XGB) & 0.10 & 0.53 & 0.01 & 0.01 & 0.24 \\
T (LR) & 0.10 & 0.53 & 0.01 & 0.01 & 0.24 \\
T (NN) & 0.07 & 0.50 & -0.00 & 0.06 & 0.24 \\
T (XGB) & 0.10 & 0.51 & 0.01 & 0.01 & 0.24 \\
TARNet & 0.07 & 0.48 & -0.00 & 0.01 & 0.21 \\
X (LR) & 0.10 & 0.53 & -0.00 & 0.01 & 0.24 \\
X (NN) & 0.08 & 0.50 & -0.02 & 0.03 & 0.24 \\
X (XGB) & 0.09 & 0.51 & 0.01 & 0.01 & 0.24 \\
\bottomrule
\end{tabular}
\caption{HILLSTROM MENS}\label{tab:real_hillstrom_mens_avg_metric_values}
\end{table}

\begin{table}[h!]
\centering
\begin{tabular}{lrrrrr}
\toprule
Method & Uplift@20 & Qini & RATE & $\Delta$ ATE & Factual \\
\midrule
CFR & 0.04 & 0.48 & -0.00 & 0.01 & 0.19 \\
Causal Forest & 0.06 & 0.57 & 0.01 & 0.01 & NaN \\
DR (LR) & 0.07 & 0.65 & 0.01 & 0.00 & 0.22 \\
DR (NN) & 0.07 & 0.62 & 0.03 & 0.01 & 0.22 \\
DR (XGB) & 0.07 & 0.65 & 0.02 & 0.00 & 0.22 \\
DragonNet & 0.07 & 0.63 & 0.02 & 0.01 & 0.22 \\
R (LR) & 0.07 & 0.65 & 0.02 & 0.00 & 0.22 \\
R (NN) & 0.07 & 0.62 & 0.02 & 0.01 & 0.20 \\
R (XGB) & 0.07 & 0.64 & 0.02 & 0.00 & 0.22 \\
S (LR) & 0.05 & 0.49 & 0.00 & 0.04 & 0.22 \\
S (NN) & 0.04 & 0.49 & -0.00 & 0.04 & 0.21 \\
S (XGB) & 0.07 & 0.64 & 0.02 & 0.01 & 0.22 \\
T (LR) & 0.06 & 0.60 & 0.01 & 0.00 & 0.22 \\
T (NN) & 0.05 & 0.53 & 0.01 & 0.06 & 0.22 \\
T (XGB) & 0.07 & 0.64 & 0.02 & 0.00 & 0.22 \\
TARNet & 0.05 & 0.50 & 0.00 & 0.01 & 0.19 \\
X (LR) & 0.07 & 0.64 & 0.03 & 0.00 & 0.22 \\
X (NN) & 0.07 & 0.58 & 0.01 & 0.03 & 0.22 \\
X (XGB) & 0.07 & 0.65 & 0.02 & 0.00 & 0.22 \\
\bottomrule
\end{tabular}
\caption{HILLSTROM WOMENS}
\label{tab:real_hillstrom_womens_avg_metric_values}
\end{table}

\begin{table}[h!]
\centering
\begin{tabular}{lrrrrr}
\toprule
Method & Uplift@20 & Qini & RATE & $\Delta$ ATE & Factual \\
\midrule
CFR & 2.28 & 0.43 & -0.22 & 0.92 & 5.63 \\
Causal Forest & 4.12 & 0.52 & 0.52 & 1.12 & NaN \\
DR (LR) & 4.07 & 0.52 & 0.76 & 1.20 & 5.38 \\
DR (NN) & 2.30 & 0.47 & -0.71 & 2.21 & 8.20 \\
DR (XGB) & 3.65 & 0.55 & 0.84 & 1.27 & 5.43 \\
DragonNet & 3.09 & 0.48 & 0.69 & 1.43 & 5.47 \\
R (LR) & 4.07 & 0.55 & 0.79 & 1.18 & 5.36 \\
R (NN) & 1.65 & 0.46 & -1.21 & 2.39 & 7.02 \\
R (XGB) & 4.88 & 0.55 & 0.82 & 1.20 & 5.49 \\
S (LR) & 3.49 & 0.51 & 0.45 & 1.19 & 5.36 \\
S (NN) & 3.57 & 0.53 & 0.71 & 1.96 & 6.52 \\
S (XGB) & 4.89 & 0.59 & 1.27 & 0.98 & 5.40 \\
T (LR) & 4.70 & 0.54 & 0.98 & 1.26 & 5.38 \\
T (NN) & 1.54 & 0.44 & 0.22 & 2.01 & 7.23 \\
T (XGB) & 3.97 & 0.53 & 0.77 & 1.24 & 5.48 \\
TARNet & 1.64 & 0.44 & -1.08 & 0.97 & 5.73 \\
X (LR) & 4.70 & 0.54 & 0.98 & 1.26 & 5.38 \\
X (NN) & 1.54 & 0.44 & 0.22 & 2.01 & 7.23 \\
X (XGB) & 3.97 & 0.53 & 0.77 & 1.24 & 5.48 \\
\bottomrule
\end{tabular}
\caption{NHEFS}
\label{tab:real_nhefs_avg_metric_values}
\end{table}

\begin{table}[h!]
\centering
\begin{tabular}{lrrrrr}
\toprule
Method & Uplift@20 & Qini & RATE & $\Delta$ ATE & Factual \\
\midrule
CFR & -4.65 & 0.00 & -0.70 & 7.95 & 242.39 \\
Causal Forest & 7.41 & 0.31 & 2.57 & 4.58 & NaN \\
DR (LR) & 4.47 & 0.32 & 2.38 & 3.77 & 275.31 \\
DR (NN) & 6.10 & 0.33 & 1.11 & 4.20 & 245.36 \\
DR (XGB) & -11.72 & -0.38 & 6.76 & 3.73 & 239.02 \\
DragonNet & 6.28 & 0.13 & 2.94 & 6.53 & 241.63 \\
R (LR) & 4.19 & 0.32 & 3.10 & 3.76 & 275.32 \\
R (NN) & 5.77 & 0.32 & 2.29 & 3.69 & 241.33 \\
R (XGB) & -11.50 & -0.39 & 3.11 & 3.59 & 239.08 \\
S (LR) & 5.86 & 0.17 & 0.30 & 3.80 & 275.32 \\
S (NN) & -1.71 & 0.10 & 2.55 & 4.06 & 240.92 \\
S (XGB) & -0.29 & -0.15 & 1.98 & 3.56 & 239.05 \\
T (LR) & 3.89 & 0.29 & 3.25 & 3.80 & 275.32 \\
T (NN) & -2.64 & 0.21 & 4.02 & 6.59 & 241.99 \\
T (XGB) & -47.41 & -0.47 & 3.62 & 3.72 & 239.29 \\
TARNet & -2.49 & 0.11 & -0.61 & 7.87 & 242.37 \\
X (LR) & 3.72 & 0.30 & 2.46 & 3.80 & 275.32 \\
X (NN) & 5.74 & 0.21 & -0.55 & 4.93 & 241.99 \\
X (XGB) & -45.63 & -0.63 & -3.13 & 3.68 & 239.29 \\
\bottomrule
\end{tabular}
\caption{RETAIL}
\label{tab:real_retail_avg_metric_values}
\end{table}

\begin{table}[h!]
\centering
\begin{tabular}{lrrrrr}
\toprule
Method & Uplift@20 & Qini & RATE & $\Delta$ ATE & Factual \\
\midrule
S (LR) & 5.29 & 862.75 & 3.00 & 1.04 & 5.16 \\
S (XGB) & 5.25 & 860.00 & 3.16 & 0.94 & 5.84 \\
T (LR) & 4.58 & 742.07 & 2.65 & 0.24 & 5.18 \\
T (XGB) & 5.20 & 855.06 & 3.18 & 0.14 & 5.83 \\
X (LR) & 5.10 & 831.08 & 2.49 & 0.17 & 5.18 \\
X (XGB) & 5.35 & 865.13 & 2.82 & 0.20 & 4.96 \\
\bottomrule
\end{tabular}
\caption{CRITEO ($10^3$)}
\label{tab:criteo_avg_metric_values}
\end{table}

\begin{table}[h!]
\centering
\begin{tabular}{lrrrrrrrr}
\toprule
Method & Uplift@20 & Qini & RATE & $\Delta$ ATE & Factual & Policy & $\Delta$ ATE (Sim) & $\sqrt{\mathrm{PEHE}}$ \\
\midrule
S (LR) & 679.62 & 468.81 & -23.23 & 2.24 & 848.37 & 87.50 & 7.30 & 1851.29 \\
S (XGB) & 2594.97 & 822.49 & 1154.78 & 7.22 & 798.53 & 86.85 & 0.86 & 1418.74 \\
T (LR) & 2031.06 & 747.28 & 555.64 & 9.92 & 833.73 & 85.31 & 2.26 & 1740.48 \\
T (XGB) & 2601.93 & 823.24 & 1157.77 & 7.14 & 798.10 & 87.02 & 0.93 & 1415.71 \\
X (LR) & 2031.09 & 747.28 & 555.64 & 9.93 & 833.73 & 85.31 & 2.26 & 1740.48 \\
X (XGB) & 2601.82 & 823.22 & 1158.16 & 7.12 & 798.10 & 86.92 & 1.03 & 1415.84 \\
\bottomrule
\end{tabular}
\caption{CRITEO ITE ($10^3$)}
\label{tab:criteo_ite_avg_metric_values}
\end{table}

\end{document}